\documentclass[10pt,twocolumn,letterpaper]{article}

\usepackage{cvpr}
\usepackage{times}
\usepackage{epsfig}
\usepackage{graphicx}
\usepackage{amsmath}
\usepackage{amssymb}

\usepackage{colortbl}
\usepackage[table]{xcolor}
\usepackage{caption}

\usepackage{amsmath,amsfonts,amssymb}

\usepackage[pagebackref=true,breaklinks=true,letterpaper=true,colorlinks,bookmarks=false]{hyperref}

\cvprfinalcopy 

\def\cvprPaperID{1295} 

\begin{document}

\title{Recursive Recurrent Nets with Attention Modeling for OCR in the Wild}

\author{Chen-Yu Lee$^{\dagger }$\\
UC San Diego\\
{\tt\small chl260@ucsd.edu}
\and
Simon Osindero\\
Flickr / Yahoo Inc.\\
{\tt\small osindero@yahoo-inc.com}
}

\maketitle

\begin{abstract}
\vspace{-2mm}
We present recursive recurrent neural networks with attention modeling (R$^2$AM) for lexicon-free optical character recognition in natural scene images. The primary advantages of the proposed method are: (1) use of recursive convolutional neural networks (CNNs), which allow for parametrically efficient and effective image feature extraction; (2) an implicitly learned character-level language model, embodied in a recurrent neural network which avoids the need to use N-grams; and (3) the use of a soft-attention mechanism, allowing the model to selectively exploit image features in a coordinated way, and allowing for end-to-end training within a standard backpropagation framework.
 
We validate our method with state-of-the-art performance on challenging benchmark datasets: Street View Text, IIIT5k, ICDAR and Synth90k. 

\end{abstract}

\let\thefootnote\relax\footnotetext{$^{\dagger }$Author to whom correspondence should be addressed.}
\let\thefootnote\relax\footnotetext{This work is supported by Flickr Vision and Machine Learning Team.}

\vspace{-3mm}
\section{Introduction}

Photo Optical Character Recognition (photo OCR), which aims to read scene text in natural images, is an essential step for a wide variety of computer vision tasks, and has enjoyed significant success in several commercial applications. These include street-sign reading for automatic navigation systems, assistive technologies for the blind (such as product-label reading), real-time text recognition and translation on mobile phones, and search/indexing the vast corpus of image and video on the web.

The field of photo OCR has been primarily focused on constrained scenarios with hand-engineered image features. (Here, constrained means that there is a fixed lexicon or dictionary and words have known length during inference.). Specifically, examples of constrained text recognition methods include region-based binarization or grouping \cite{chen2004detecting, kita2010binarization, neumann2012real}, pictorial structures with HOG features \cite{wang2010word, wang11}, integer programming with SIFT descriptor \cite{smith2011enforcing}, Conditional Random Fields (CRFs) with HOG features \cite{mishra2012cvpr, Mishra12, shi2013scene}, Markov models with binary and connected component features \cite{weinman2014toward}. 
Some early attempts \cite{lee2014, yao2014, gordo2015} try to learn local mid-level representation on top of the hand-crafted features, and some methods in \cite{wang2012, jader2014, Jaderberg14c} incorporate deep convolutional neural networks (CNNs) \cite{lecun1998gradient, hinton2006fast} for a better image feature extraction. These methods work very well when candidate ground-truth word strings are known at testing stage, but do not generalize to words that are not present in the list of a lexicon at all.

A recent advance in the state-of-the-art that moves beyond this constrained setting was presented by Jaderberg \etal in \cite{jader2015b}. The authors report results in the unconstrained setting by constructing two sets of CNNs --  one for modeling character sequences and one for N-gram language statistics --  followed by a CRF graphical model to combine their activations. This method achieved great success and set a new standard in photo OCR field. However, despite these successes, the system in \cite{jader2015b} does have some drawbacks. For instance, the use of two different CNNs incurs a relatively large memory and computation cost. Furthermore, the manually defined N-gram CNN model has a large number of output nodes ($10k$ output units for N = 4), which increases the training complexity -- requiring an incremental training procedure and heuristic gradient rescaling based on N-gram frequencies.

Inspired by \cite{jader2015b}, we continue to focus our efforts on the unconstrained scene text recognition task, and we develop a recursive recurrent neural networks with attention modeling (R$^2$AM) system that directly performs image to sequence (word strings) learning, delivering improvements over their work. The three main contributions of the work presented in this paper are:

\noindent(1) Recursive CNNs with weight-sharing, for more effective image feature extraction than a ``vanilla'' CNN under the same parametric capacity. 

\noindent(2) Recurrent neural networks (RNNs) atop of extracted image features from the aforementioned recursive CNNs, to perform implicit learning of character-level language model. RNNs can automatically learn the sequential dynamics of characters that are naturally present in word strings from the training data without the need of manually defining N-grams from a dictionary.

\noindent(3) A sequential attention-based modeling mechanism that performs ``soft'' deterministic image feature selection as the character sequence is being read, and that can be trained end-to-end within the standard backpropagation.

We pursue extensive experimental validation on challenging benchmark datasets: Street View Text, IIIT5k, ICDAR and Synth90k. We also provide a detailed ablation study by examining the effectiveness of each of the proposed components. Our proposed network architecture achieves the new state-of-the-art results and significantly outperforms the previous best reported results for unconstrained text recognition \cite{jader2015b}; \ie we observe an absolute accuracy improvement of 9\% on Street View Text and 8.2\% on ICDAR 2013.

\section{Methodology}
\begin{figure}[t]
\begin{center}
\includegraphics[width=0.92\linewidth]{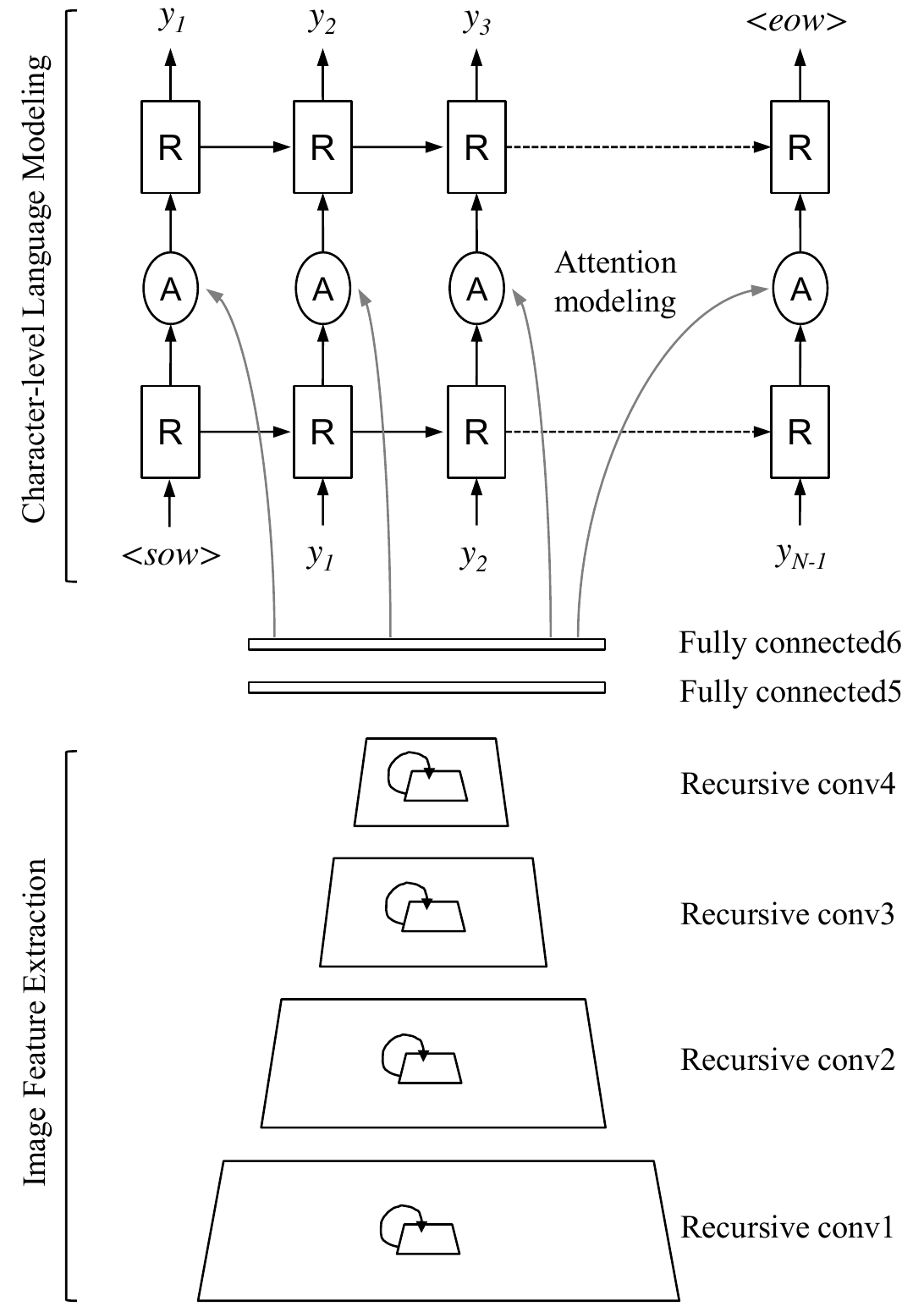}
\end{center}
\vspace{-6mm}
   \caption{Recursive recurrent nets with attention modeling (R$^2$AM) approach: the model first passes input images through recursive convolutional layers to extract encoded image features, and then decodes them to output characters by recurrent neural networks with implicitly learned character-level language statistics. 
Attention-based mechanism performs soft feature selection for better image feature usage. }
\vspace{-3mm}
\label{fig:pipeline}
\end{figure}

In this paper, we focus on the scene text recognition task,  predicting all characters from a cropped image of single word. We refer to the cropped word region as an input image in the rest of the paper. The current section describes related literatures and the proposed architecture: Recursive Recurrent Nets with Attention Modeling (R$^2$AM). Figure \ref{fig:pipeline} shows our overall system architecture.

\subsection{Character sequence model review}
\label{baseCNN}
Many text recognition methods focus on capturing individual characters of a word as the first step in the system pipeline, and then apply statistical language models or visual structure prediction to refine/prune-out misclassified characters as in \cite{wang11, wang2012, mishra2012cvpr, bissa2013, shi2013scene, lee2014, yao2014}. 
However, there are significant challenges since each character is intimately positioned with respect to others within the same word, and therefore classic character recognition components need to deal with a large amount of inter-class and intra-class confusion -- this is well illustrated by Figure 3 from \cite{mishra2012cvpr}. Even in  sophisticated word recognition systems incorporating higher-order language priors based on CRFs or Markov models, the overall system performance is still largely dominated by the capability of the first step of the system pipeline: the character recognition component.

Goodfellow \etal \cite{goodfellow2013multi} first used a CNN with multiple position-sensitive character classifiers for street number recognition. Recently, Jaderberg \etal in \cite{jader2015, jader2015b} proposed character sequence model that directly encodes the character at each position in the word using deep CNNs and so predicts the sequence of characters in an image region. This approach largely overcomes the aforementioned issues by directly modeling the natural spacing and overlapping patterns in scene characters that can not readily be leveraged by sliding window based character recognition methods. For details of this character sequence model please refer to \cite{jader2015b}. We refer to this baseline method as Base CNN (and labeled in Figure \ref{fig:5methods} as Base CNN) in the rest of the paper. Our proposed system is built upon this Base CNN model; we describe our extension of novel image encoding in Section \ref{spaceR}, character-level language modeling in Section \ref{timeR}, and attention-based mechanism in Section \ref{attenR}.

\subsection{Recursive CNNs for image feature extraction}
\label{spaceR}
\subsubsection{Recursive convolutional layers}
One key to the great success of the aforementioned character sequence model is the ability to capture contextual dependencies during character prediction by employing multiple convolutional layers that operate on the whole input image. 

One possible way to improve upon this Base CNN model to enable even longer range contextual dependencies for character prediction would be to consider using a larger kernel size for each convolutional layer or a deeper network, increasing the corresponding receptive field size. However, this approach induces more parameters and a higher model complexity, thereby leading to potential training and generalization issues.

Another way to expand longer data dependencies while controlling the model capacity is to make the Base CNN network recursive or recurrent as suggested in \cite{pinheiro2013, eigen2013understanding, liang2015recurrent}. By using recursive or recurrent convolutional layers, the network architecture can be arbitrary deep without significantly increasing the total number of parameters by reusing the same convolutional weight matrix multiple times at each layer.

We now describe the recursive CNNs used in our approach: the instance of the recursive convolutional layer at time step $t$ (where $t \geq 0$) is fed with an input image/feature response as:
\begin{equation}
\small
\begin{aligned}
\label{eq:2}
    h_{i,j,k}(t)=
   \begin{cases}
    \sigma((\mathbf{w}_{k}^{hh})^T \: \mathbf{x}_{i,j} + b_{k}) & \text{ at $t=0$ } \\
    \sigma((\mathbf{w}_{k}^{hh})^T \: \mathbf{h}_{i,j}(t-1) + b_{k}) & \text{ at $t > 0$ } \\
   \end{cases} 
\end{aligned}
\end{equation}
where $\mathbf{h}_{i,j}(t-1)$ and $\mathbf{x}_{i,j}$ denote the vectorized feed-forward and input patches centered at $(i,j)$ of feature maps, respectively. $\mathbf{w}_{k}^{hh}$ is the vectorized feed-forward weight for output channel $k$. $b_{k}$ is the bias for output channel $k$. $\sigma$ is a deterministic non-linear transition function. 

Recursive CNNs increase the depth of traditional CNNs under the same parametric capacity, and also produce much more compact feature response than CNNs. In a slightly different interpretation of this architecture, the recursive interactions can also be seen as implementing a form of ``lateral connectivity'' within a feature map, allowing the representation at a given layer to better capture higher order dependencies. For a longer discussion on recursive/recurrent convolutional layers, see \cite{liang2015recurrent} for details.

\subsubsection{Untying in recursive convolutional layers}
We have seen the definition and potential gain of recursive convolutional layers in the previous section. However, the formulation in Eqn.~\ref{eq:2} restricts all weights $\mathbf{w}_{k}^{hh}$ to share the same internal values -- they are ``tied" together. One consequence of this tying is that the number of channels will be identical across all of the layers due to the fact that the shared weights $\mathbf{w}_{k}^{hh}$ always project the incoming feature maps to the same dimension (width$\times$height$\times$number of channels) of output feature maps. This strongly contrasts with the common practice in CNNs of varying the number of channels to control the amount of computation performed and the spectrum of different feature types. (For example the popular and successful VGGNet \cite{simonyan2014very}, in which the number of channels increases like \{64, 128, 256, 512\} as the spatial extent of the convolutional layers decreases according to \{224, 112, 56, 28\}.)

\begin{figure}[t]
\begin{center}
\includegraphics[width=0.85\linewidth]{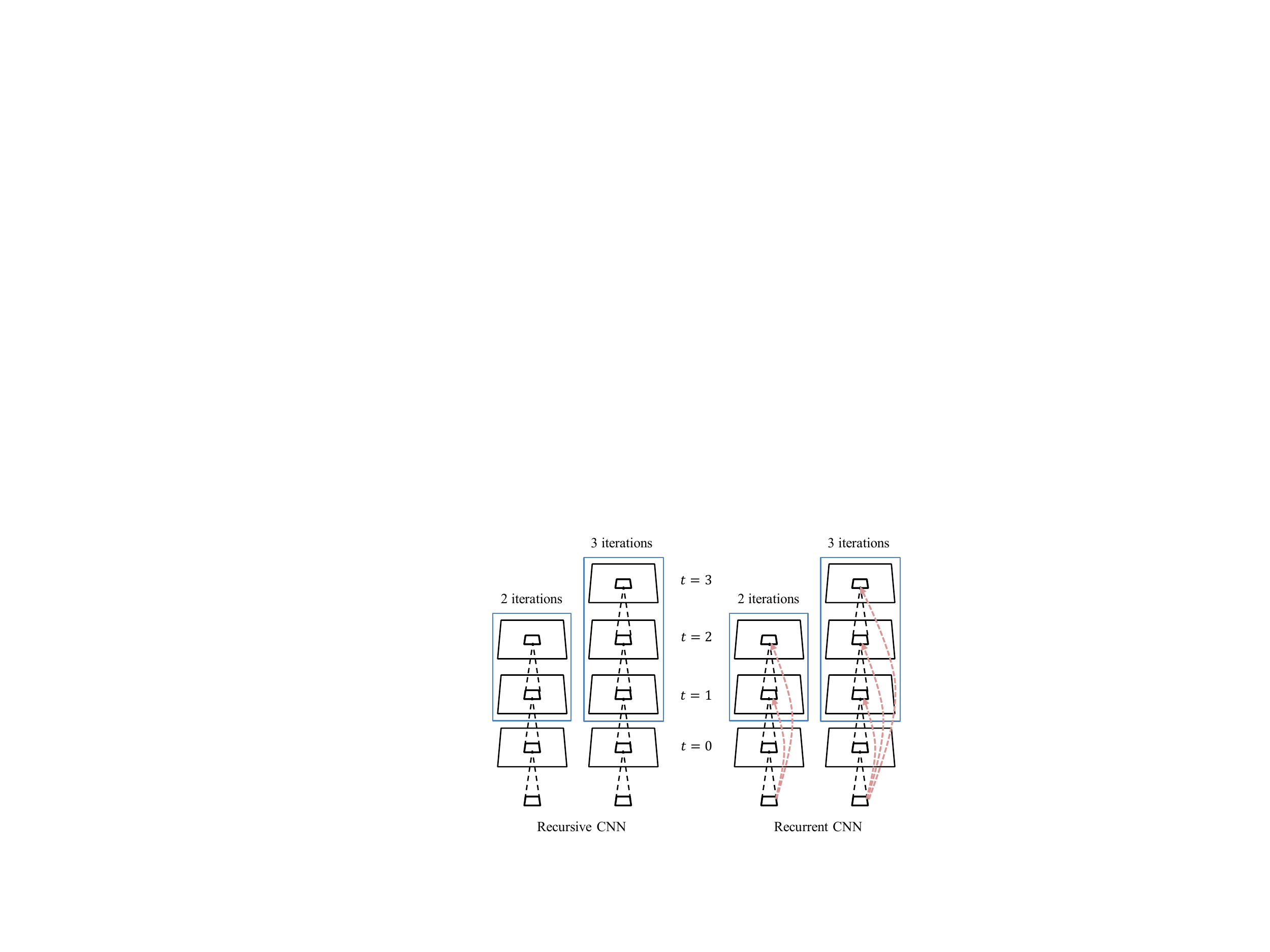}
\end{center}
    \vspace{-6mm}
   \caption{Illustration of the proposed untied recursive and recurrent convolutional layers. We untie the first feed-forward weight at time step $t=0$ and the rest feed-forward weights at $t \geq 1$. The layers inside the blue box have tied (shared) weights. }
\label{fig:iteration}
\end{figure}

In this work we propose to use  an ``untied'' variant of recursive convolutional layers that distinguishes between initial inter-layer feed-forward weight $\mathbf{w}_{\text{untied},k}^{hh}$ and the following intra-layer recursive weights $\mathbf{w}_{\text{tied},k}^{hh}$. This allows the network to have different numbers of channels at different layers, and also allows the recursive weights to specialize more freely. 

By untying the feed-forward weights at time step $t=0$,   Eqn. \ref{eq:2} becomes:
\begin{equation}
\small
\begin{aligned}
    h_{i,j,k}(t)=
   \begin{cases}
  \sigma ((\mathbf{w}_{\text{untied},k}^{hh})^T \:  \mathbf{x}_{i,j} + b_{k}) & \text{ at $t=0$ } \\
   \sigma ((\mathbf{w}_{\text{tied},k}^{hh})^T \: \mathbf{h}_{i,j}(t-1) + b_{k}) & \text{ at $t > 0$}
   \end{cases} 
\end{aligned}
\end{equation}
In doing so, the number of channels for any recursive convolutional layer can be adjusted by the untied weight $\mathbf{w}_{\text{untied},k}^{hh}$, controlling the overall computational cost. We can use the same logic here to untie recurrent convolutional layer \cite{liang2015recurrent}. Please see Figure \ref{fig:iteration} for an illustration. 

In the experiment section we observed that both recursive and recurrent versions of Base CNN model significantly improve the performance on many recent standard benchmarks such as Synth90k, SVT, and ICDAR13. Please refer the details in Section \ref{ablation}. We further found that recursive version consistently outperforms recurrent version in all the tasks that we explored, which is in line with findings in other recent literature \cite{socher2012, irsoy2014} that recursive structures can learn compositional features and part interactions effectively so injecting input $\mathbf{x}_{i,j}$ multiple times at each time step is not necessary for obtaining high performance. It is also possible that recursive models are forced to more effectively use their tied weights relative to the recurrent model, since there is no option for information to ``short-cut'' as there is in the recurrent architecture. For this reason, we choose the recursive version of Base CNN model for our overall system pipeline as shown in the bottom part of Figure \ref{fig:pipeline}.

\begin{figure*}[ht]
\begin{center}
\includegraphics[width=0.8\linewidth]{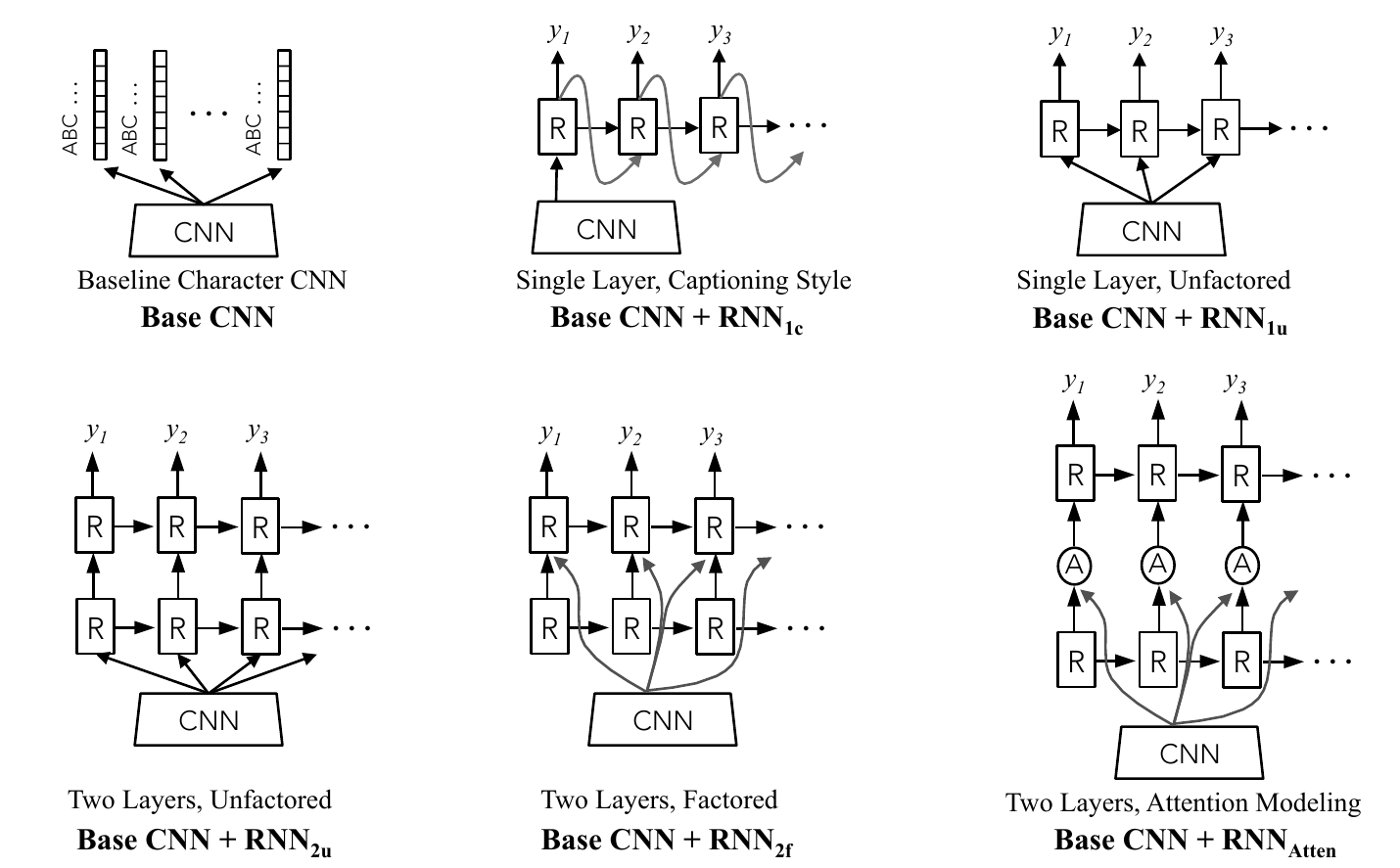}
\end{center}
    \vspace{-6mm}
   \caption{Five variations of the recurrent in time architecture that we experimentally evaluate for photo OCR task. We explore the image captioning style RNNs, the effect of depth in the RNN stack, the effect of factorization of the modalities (also explored in \cite{kiros2014, donahue2015}), and the effect of attention modeling. }
\label{fig:5methods}
\vspace{-1mm}
\end{figure*}

\subsection{RNNs for character-level language modeling}
\label{timeR}
The proposed untied recursive character sequence model in Section \ref{spaceR} can already serve as an end-to-end trainable photo OCR system that significantly outperforms sophisticated structured learning method in \cite{jader2015b} by a large margin (7.2\% on SVT and 6.7\% on ICDAR13 absolute improvement as shown in the experiment section). Nonetheless, we observed that the Base CNN model (either plain CNNs, recursive CNNs, or recurrent CNNs) trains each character position independently by using multiple loss functions. We are then motivated to ask whether we can allow some sort of interaction between each character position and exploit the underlying character-level language statistics.

The most common approach is to add some kind of graphical models (\eg CRFs) on top of the output prediction of each character position as in \cite{Mishra12, mishra2012cvpr, novik2012}. However, these methods need to compute unary and higher-order terms for all candidate characters, and can require expensive computation during inference stage.
Another way to access such character-level language information is to directly model all possibilities using a CNN -- as in the bag-of-N-grams component of \cite{jader2015b}. Such a CNN model requires pre-defined N-grams from a dictionary and uses a huge number of output nodes in which each node represents an element in N-gram combinations (\eg $10k$ output nodes for $N=4$ in \cite{jader2015b}). In order to jointly train a whole range of N-grams in this CNN model using backpropagation, the method in \cite{jader2015b} rescales the gradients for each N-gram class by the inverse frequency of its appearance in the training word corpus because some of the N-gram classes barely occur in the training dataset, even with the recently released Synth90k \cite{Jaderberg14c} dataset that has more than $7$ million training samples.

In contrast to the above methods, we propose to use recurrent neural networks (RNNs) \cite{werbos1988generalization, schmidhuber2015deep, rumelhart1985learning} to model the character-level statistics of text. Recurrent neural networks and its variant Long Short-Term Memory (LSTM) \cite{hochreiter1997long} have recently experienced a renaissance and are extremely effective models when dealing with sequential data, such as handwriting recognition, machine translation, speech recognition, and image captioning. 
Recognizing characters in images can be essentially considered as a task of solving sequential dynamics and learning mappings from pixel intensities to natural character-level vectors. Specifically, our model takes a single image and generate a sequence of 1-of-$K$ encoded characters.
\begin{equation}
\small
\mathcal{Y} = \{\mathbf{y}_1,\mathbf{y}_2,...,\mathbf{y}_N \}, \: \mathbf{y}_t \in \mathbb{R}^K
\end{equation}
where $K$ is the size of the possible characters and $N$ is the length of the word. 

We propose the use of an RNN that produces a word string by generating one character at every time step conditioned on image feature $\boldsymbol{\mathcal{I}}$, the previous hidden state $\mathbf{h}_{t-1}$ and the input $\mathbf{x}_{t}$ using the recurrence equations:
\begin{equation}
\small
\begin{aligned}
&\mathbf{h}_t = \sigma (W_{xh} \mathbf{x}_{t} + W_{hh} \mathbf{h}_{t-1} + \mathbf{b}_h) \\
&\mathbf{y}_t = \sigma (W_{hy} \mathbf{h}_{t} + \mathbf{b}_y) 
\end{aligned}
\end{equation}
where $\sigma$ is an element-wise non-linear transition function, $\mathbf{h}_{t} \in \mathbb{R}^M$ is the hidden state with $M$ units, and the input $\mathbf{x}_{t}$ can be encoded image feature $\boldsymbol{\mathcal{I}}$ or previously generated character $\mathbf{y}_{t-1} $, depending on the RNN structure used. The encoded image feature $\boldsymbol{\mathcal{I}}$ is extracted from the last fully-connected layer of the a CNN model. This CNN models can be either plain CNN, recurrent CNN, or recursive CNN. We will show how different CNN models perform in the experiment section.

There are potentially many ways to feed an image feature $\boldsymbol{\mathcal{I}}$ to a RNN, and the RNN itself can also have many different structures. In this paper, we empirically explore a range of settings. Figure \ref{fig:5methods} demonstrates the base CNN model and five RNN variants that we explored. We detail four variants in this section and will explain the last variant (including attention modeling) in the next section:

\noindent \textbf{Base CNN:}
Baseline character sequence CNN trained with multiple loss functions where each loss function focuses on one character position as described in Section \ref{baseCNN}.

\noindent \textbf{Base CNN + RNN$_{\textbf{1c}}$:}
A single-layer RNN inspired from image-captioning work \cite{vinyals2015}. The extracted image feature $\boldsymbol{\mathcal{I}}$ is sent to RNN only at the first time step. The predicted character $y_{t-1}$ of RNN at time $t-1$ is fed to the RNN at time $t$ until we obtain an end-of-word (EOW) label. This variant serves as an good sanity check and helps us validate the capability of our RNN to perform character-level language modeling given an initial CNN representation.

\noindent \textbf{Base CNN + RNN$_{\textbf{1u}}$:}
An unfactored single-layer RNN receiving image feature $\boldsymbol{\mathcal{I}}$ at every time step -- therefore the character predictions are conditioned on both image feature and previous hidden state at all time.

\noindent \textbf{Base CNN + RNN$_{\textbf{2u}}$:}
An unfactored two-layer RNN using two stacks of RNNs. This model has a deeper structure at each time step. This variant also has access to image feature at every time step.

\noindent \textbf{Base CNN + RNN$_{\textbf{2f}}$:}
A factored two-layer RNN that uses two stacks of RNNs. This variant only has access to the image features at the second layer RNN. In this way we force the first stack of RNN to focus on character-level language modeling and force the second stack of RNN to focus on combining language statistics and image feature.

As noted previously, all the RNN variants that we explored implicitly perform character-level language modeling and benefit from not being constrained to pre-defined N-gram sequences. For instance, Karpathy \etal \cite{karpathy2015visualizing} demonstrate that RNN-based methods consistently outperforms N-gram models at character-level text prediction where N is as large as 20. In the experiment section we will show the error analysis with and without the proposed RNNs.

\subsection{Attention modeling}
\label{attenR}
Attention-based mechanisms can allow the model to focus on the most important segments of incoming features, as well as potentially adding a degree of interpretability  \cite{bahdanau2014neural, xu2015show}. There are generally two categories of attention-based image understanding: hard-attention and soft-attention. Hard-attention models learn to choose a series of discrete glimpse locations, and can be challenging to train since the loss gradients are typically intractable. In this work we choose a soft-attention model, which can be trained end-to-end with standard backpropagation. 

We now describe our attention modeling function illustrated in Figure \ref{fig:5methods} as \textbf{Base CNN + RNN$_{\textbf{Atten}}$}. At every output step $t$, the attention function (denoted as a letter A in the figure) computes an energy vector $\boldsymbol{\tau}_t$ conditioned on the image feature $\boldsymbol{\mathcal{I}}$ and the output of the first stack RNN $\mathbf{s}_t$:
\begin{equation}
\small
\boldsymbol{\tau}_t  = f_{\text{attention}}(\boldsymbol{\mathcal{I}},  \mathbf{s}_t) = \tanh ( \varphi (\boldsymbol{\mathcal{I}}) + \psi (\mathbf{s}_t) ) 
\end{equation}
where $\varphi$ and $\psi$ can be multilayer perceptrons or simple weight matrices that project both $\boldsymbol{\mathcal{I}}$ and $\mathbf{s}_t$ to the same space. Then the context vector $\mathbf{c}_t$ is computed as weighted image feature based on the energy coefficients $\alpha_t$ at time step $t$:
\begin{equation}
\small
\begin{aligned}
\alpha_{td} & = \frac{\exp(\tau_{td})}{\sum_{d=1}^{D}\exp(\tau_{td})} \\
\mathbf{c}_t  & = \boldsymbol{\alpha}_t \circ \boldsymbol{\boldsymbol{\mathcal{I}}} 
\end{aligned}
\end{equation}
where $\circ$ is the Hadamard product. This mechanism generates a set of positive weights $\alpha_{td}$ which can be understood as the relative importance to give to location $d$ in fusing the image feature $\boldsymbol{\mathcal{I}}$, and the computed context vector $\mathbf{c}_t$ is then sent to the second stack of RNN for final output prediction. 

\section{Experiments}
\label{experiment}
\subsection{Datasets}
We evaluate the proposed Recursive Recurrent Nets with Attention Modeling (R$^2$AM) framework on five standard benchmark datasets: ICDAR 2003, ICDAR 2013, Street View Text, IIIT5k and Synth90k. 

\noindent \textbf{ICDAR 2003} \cite{lucas2003icdar} contains 251 full scene images and 860 cropped images of the words. Even though the focus of this paper is unconstrained text recognition, we nonetheless provide constrained text recognition results for the ease of comparison. The per-image 50 word lexicons defined by Wang \etal \cite{wang11} are referred to as IC03-50 and the lexicon of all test words (563 words) is referred as IC03-Full.

\noindent \textbf{ICDAR 2013} \cite{karatzas2013icdar} contains 1015 cropped word images from natural scene images and is referred as IC13.

\noindent \textbf{Street View Text} \cite{wang11} contains 647 cropped word images from Google Street View. The per-image 50 word lexicons defined by Wang \etal \cite{wang11} are referred to as SVT-50 and the lexicon of all test words (4282 words) is referred as SVT-Full.

\noindent \textbf{IIIT5k} \cite{Mishra12} contains 3000 cropped word images downloaded from Google image search engine. Each image has a lexicon of 50 word (IIIT5k-50) and a lexicon of 1k word (IIIT5k-1k).

\noindent \textbf{Synth90k} \cite{Jaderberg14c} contains synthetically generated word images. The dataset contains around 7 million training images, 900k validation images, and 900k test images.

We follow the setting in Jaderberg \etal \cite{jader2015b} to prepare training and test sets that our method is trained purely on the Synth90k training set and all parameters are selected via validation set. We do not use the validation data to retrain our model. We also follow the evaluation protocol in Wang \etal \cite{wang11} that performs recognition on the words containing only alphanumeric characters (0-9 and A-Z) and at least three characters.

\subsection{Implementation details}
The network architecture for our Base CNN model is shown in Table \ref{tab:baseCNN}. It has 8 convolutional layer with 64, 64, 128, 128, 256, 256, 512 and 512 channels, and each convolutional layer uses kernel with a  3 $\times$ 3 spatial extent. Convolutions are performed with stride 1, zero padding, and ReLU activation function. 2 $\times$ 2 max pooling follows the second, fourth, and sixth convolutional layers. The two fully connected layers have 4096 units. The input is a resized 32 $\times$ 100 gray scale image.

We now provide details for the network structures of the proposed untied recursive CNNs in Table \ref{tab:baseCNN}. Notice that each of the even number convolutional layer (conv2, conv4, conv6 or conv8) use its own shared weight matrix that has exactly the same input and output dimensionality, and so projects feature maps to the same space multiple times within one recursive convolutional layer under the same parametric capacity as Base CNN model. 

For the character-level language modeling, we use RNNs with 1024 hidden units equipped with hyperbolic tangent activation function. Our overall system pipeline is shown in Figure \ref{fig:pipeline}.

We apply backpropagation through time (BPTT) algorithm to train the models with 256 batch size SGD and 0.5 dropout rate. Initial learning rate is 0.002 and decreased by a factor of 5 as validation errors stop decreasing for 2 epochs. All variants use the same scheme with 30 total epochs determined based on the validation set. We apply gradient clipping at the magnitude of 10, and find it with in place weight decay did not add extra performance gains. All initial weights are sampled from Gaussian distribution with 0.01 standard deviation. We implemented the system in the open source deep learning framework Caffe \cite{jia2014caffe}. The average inference time per image is 2.2 ms on single Nvidia Titan X GPU for the overall system framework.

\begin{table}
\footnotesize
\begin{center}
\begin{tabular}{|l|c|c|c|c|}
\hline
Method                              & Synth90k         & SVT       & ICDAR13   \\
\hline\hline
CHAR \cite{jader2015b}      &  87.3          & 68.0          & 79.5          \\
Base CNN                            &  91.9            & 75.1      & 85.7      \\
\hline
\hline
Recurrent CNN (2 iter)              &  92.6            & 75.8      & 86.1      \\
Recurrent CNN (3 iter)              &  93.5            & 76.9      & 87.4      \\
\hline
\hline
Recursive CNN (2 iter)              &  93.3            & 77.1      & 87.3      \\
Recursive CNN (3 iter)              &  \textbf{94.2}   & \textbf{78.9} & \textbf{88.5}  \\
\hline
\end{tabular}
\end{center}
\vspace{-6mm}
\caption{Unconstrained (lexicon-free) text recognition accuracies on recent benchmarks. See Figure \ref{fig:iteration} for diagrams of these architectures. The results indicate that ``iteration'' is important to both recurrent and recursive CNNs. Recursive CNNs outperform recurrent CNNs on all three datasets.}
\label{tab: recurrent_recursive}
\end{table}

\begin{table}
\footnotesize
\begin{center}
\begin{tabular}{|l|c|c|c|c|}
\hline
Method                              & Synth90k          & SVT           & ICDAR13     \\
\hline\hline
CHAR \cite{jader2015b}      &  87.3          & 68.0          & 79.5          \\
Base CNN                            &  91.9            & 75.1          & 85.7         \\
Base CNN + RNN$_{1c}$               &  93.4            & 76.2          & 86.4         \\
Base CNN + RNN$_{1u}$               &  93.5            & 76.9          & 87.2         \\
Base CNN + RNN$_{2u}$               &  93.7            & 77.9          & 87.6         \\
Base CNN + RNN$_{2f}$               &  94.0            & 78.8          & 88.0          \\
Base CNN + RNN$_{\text{Atten}}$     &  \textbf{94.3}   & \textbf{79.1} & \textbf{88.9} \\
\hline
\end{tabular}
\end{center}
\vspace{-6mm}
\caption{Unconstrained (lexicon-free) text recognition accuracies on recent benchmarks. See Figure \ref{fig:5methods} for diagrams of these architectures. The results indicate that the simply uses image captioning style method can already boost performance, while factored RNNs with attention modeling overall achieves the best results.}
\label{tab: recurrent_language}
\end{table}

\begin{table}
\footnotesize
\begin{center}
\begin{tabular}{|l|c|c|c|c|}
\hline
Method                      & Synth90k       & SVT           & ICDAR13       \\
\hline\hline
CHAR \cite{jader2015b}      &  87.3          & 68.0          & 79.5          \\
JOINT \cite{jader2015b}     &  91.0          & 71.7          & 81.8          \\
R$^2$AM (ours) &  \textbf{95.3} & \textbf{80.7} & \textbf{90.0} \\
\hline
\end{tabular}
\end{center}
\vspace{-6mm}
\caption{Unconstrained (lexicon-free) text recognition accuracies on recent benchmarks. Our combined model R$^2$AM (Recursive CNN + RNN$_{\text{Atten}}$) significantly outperforms previous state-of-the-art methods in \cite{jader2015b}.}
\label{tab:unconstraint_results}
\end{table}

\begin{table*}
\footnotesize
\begin{center}
\begin{tabular}{|l|c|>{\columncolor[gray]{0.98}}c|c|c|>{\columncolor[gray]{0.98}}c|c|c|>{\columncolor[gray]{0.98}}c|>{\columncolor[gray]{0.98}}c|}
\hline
Method                                          & SVT-50 & SVT    & IIIT5k-50 & IIIT5k-1k & IIIT5k  & IC03-50 & IC03-Full & IC03 & IC13  \\
\hline\hline                                                                                                                                  
Baseline ABBYY \cite{wang11}                    & 35.0   & -      & 24.3      & -         & -       &  56.0   & 55.0      & -    & -     \\
Wang \etal \cite{wang11}                        & 57.0   & -      & -         & -         & -       &  76.0   & 62.0      & -    & -     \\
Mishra \etal \cite{Mishra12}                    & 73.2   & -      & -         & -         & -       &  81.8   & 67.8      & -    & -     \\
Novikova \etal \cite{novik2012}                 & 72.9   & -      & 64.1      & 57.5      & -       &  82.8   & -         & -    & -     \\
Wang \etal \cite{wang2012}                      & 70.0   & -      & -         & -         & -       &  90.0   & 84.0      & -    & -     \\
Bissacco \etal \cite{bissa2013}                 & 90.4   & 78.0   & -         & -         & -       &  -      & -         & -    & 87.6  \\
Goel \etal \cite{goel2013}                      & 77.3   & -      & -         & -         & -       &  89.7   & -         & -    & -     \\
Alsharif and Pineau \cite{alsharif2013}         & 74.3   & -      & -         & -         & -       &  93.1   & 88.6      & -    & -     \\
Almaz{\'a}n \etal \cite{almazan2014}            & 89.2   & -      & 91.2      & 82.1      & -       &  -      & -         & -    & -     \\
Lee \etal \cite{lee2014}                        & 80.0   & -      & -         & -         & -       &  88.0   & 76.0      & -    & -     \\
Yao \etal \cite{yao2014}                        & 75.9   & -      & 80.2      & 69.3      & -       &  88.5   & 80.3      & -    & -     \\
Rodriguez-Serrano \etal \cite{rodriguez2014}    & 70.0   & -      & 76.1      & 57.4      & -       &  -      & -         & -    & -     \\
Jaderberg \etal \cite{jader2014}                & 86.1   & -      & -         & -         & -       &  96.2   & 91.5      & -    & -     \\
Su and Lu \etal \cite{su2015}                   & 83.0   & -      & -         & -         & -       &  92.0   & 82.0      & -    & -     \\
Gordo \cite{gordo2015}                          & 90.7   & -      & 93.3      & 86.6      & -       &  -      & -         & -    & -     \\   
*DICT Jaderberg \etal \cite{jader2015}          & 95.4   & 80.7   & 97.1      & 92.7      & -       &  98.7   & 98.6      & 93.1 & 90.8  \\
\hline                                                                                                                                      
\hline
Jaderberg \etal \cite{jader2015b}               & 93.2   & 71.7   & 95.5      & 89.6      & -       & 97.8    & \textbf{97.0}  & \textbf{89.6} & 81.8  \\
R$^2$AM (ours)          & \textbf{96.3}  & \textbf{80.7}  & \textbf{96.8} & \textbf{94.4} & \textbf{78.4}    & \textbf{97.9}    & \textbf{97.0} & 88.7 & \textbf{90.0}  \\
\hline
\end{tabular}
\end{center}
\vspace{-5mm}
\caption{Scene text recognition accuracies (\%). ``50'', ``1k'' and ``Full'' denote the lexicon size used for constrained text recognition defined in \cite{wang11}. The last two rows list methods that are capable of performing unconstrained text recognition (lexicon-free). Our proposed R$^2$AM method significantly outperforms previous best unconstrained text recognition method \cite{jader2015b} in most of the cases (bold numbers), especially on the recent released datasets such as SVT, IIIT5k, IC13. *DICT \cite{jader2015} is not lexicon-free due to incorporating ground-truth labels during training.}
\vspace{-1mm}
\label{tab:final_table}
\end{table*}

\subsection{Ablation study}
\label{ablation}
In this section we empirically investigate the contributions made by three key components in the proposed method, namely: recursive CNNs for image encoding, RNNs for character-level language modeling, and attention-based mechanism for better image feature usage. 

In an effort to decouple the performance improvement that is due to architectural variations from that which might simply come from having more parameters, we first gradually increase the depth of the baseline CHAR model in \cite{jader2015b} from 5 conv layers until we reach the performance plateau at 8 conv layers shown as Base CNN in Table \ref{tab:baseCNN}. Having observed this plateau, we did not explore even deeper networks such as  (16 or 19 conv layer) VGGNet \cite{simonyan2014very}.  The bar chart in Table \ref{tab:baseCNN} shows the corresponding performance for networks with different depths on Synth90k dataset.

   \vspace{-3mm}
\subsubsection{Recursive and recurrent convolutional layers}
Table \ref{tab: recurrent_recursive} shows the effectiveness of the proposed untied recursive and recurrent CNNs over Base CNN model on unconstrained text recognition tasks. We observed that more iterations in both proposed methods led to higher accuracies on all datatsets evaluated, and the improvement essentially came from the same parametric capacity since the weights of convolutional layer are shared as shown in Figure \ref{fig:iteration} and Table \ref{tab:baseCNN}. The lateral interactions between these shared weights allow for broader receptive fields and competition between representational units in the same ``layer''. In addition, we consistently found that recursive version outperforms recurrent version. It might because the recursive convolutional layer can prevent error signals directly backpropagate back via recurrent connections in the recurrent convolutional layer, which is especially true for convolutional operation since the input signal remains unchanged. Therefore, we choose the recursive CNNs for our final system architecture. The architectural variants we study differ from Residual Networks (ResN) \cite{he2015deep} in that our recurrent model receives the same bottom-up input at each stage of recurrency, whereas in ResN's the identity connections skip layers.

   \vspace{-3mm}
\subsubsection{Character-level language modeling} 
In Table \ref{tab: recurrent_language}, we report unconstrained text recognition results for each of the architectural variants of RNNs in Figure \ref{fig:5methods}. We observed an immediate performance boost by using any kind of the proposed RNN variants atop the Base CNN network which has already hit its performance plateau. RNN$_{1c}$ serves as a good sanity check module because the image features from the Base CNN are only fed to the RNN at the first time step, and then RNN$_{1c}$ is able to predict the first and the following character correctly based on the previously predicted character and the hidden state information. 

The comparison of RNN$_{1u}$ and RNN$_{1c}$ results indicates that feeding image feature from Base CNN to a RNN at every time step can further improve the performance, as RNN$_{1u}$ has access not only to the previously predicted character and hidden state information, but also the raw image feature during inference. (This architecture presumably also allows the RNN to expend more capacity on sequence modeling, since it no longer needs to retain the image feature information in its hidden state.) In addition, based on the observation that RNN$_{2f}$ outperforms RNN$_{2u}$, we note that factorization seems to be a more effective architecture than the unfactored models. This might because that the encoded image features can only be accessed by the second stack RNN, and this therefore allows/forces the first stack RNN to focus on modeling character-level statistics (and/or attentional processes).

The proposed RNN$_{\text{Atten}}$ uses an attention-based mechanism that learns a set of weight matrices to rescale image features and perform soft attention modeling before feeding the image feature to the top-level RNN. This architecture was observed to give the best performance across all five of the RNN variants explored.

Thus, our final network architecture contains the aforementioned recursive CNNs for image feature extraction and the RNN$_{\text{Atten}}$ for character-level language modeling with attention as shown in Figure \ref{fig:pipeline}.

\begin{figure*}[t]
\begin{center}
\includegraphics[width=0.9\linewidth]{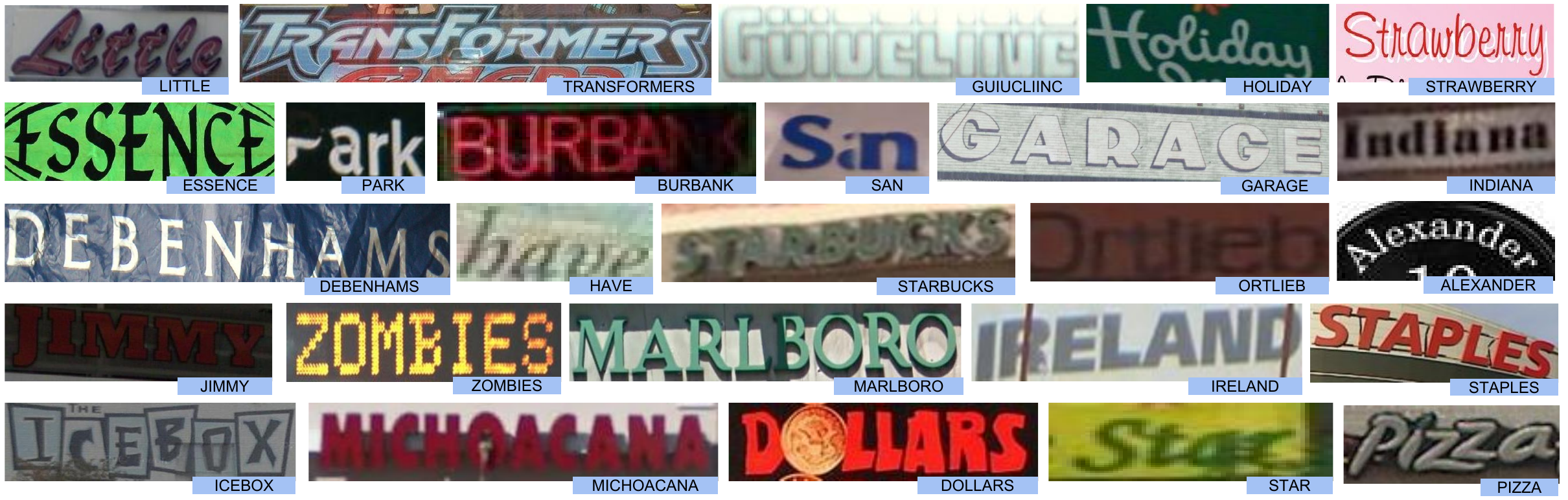}
\end{center}
\vspace{-6mm}
   \caption{Lexicon-free scene text recognition results by the proposed Recursive Recurrent Nets with Attention Modeling (R$^2$AM) framework on examples from SVT, ICDAR and IIIT5k datasets. R$^2$AM is able to recognize words with low contrast, significant transform and clutter background. It is also able to recover missing/occluded characters (\eg PARK, BURBANK, SAN and STAR) by the implicitly learned language model.  }
   \vspace{-3mm}
\label{fig:examples}
\end{figure*}

At this point it is also worth mentioning that we have  explored backward RNNs and bidirectional RNNs for character-level language modeling as well, however neither of these extensions delivered further improvements. This is in contrast to the observations in Graves \etal \cite{graves2009novel}. Perhaps this is because our work focuses on predicting characters in scene images that  contain around 8 characters on average, while method in \cite{graves2009novel} focuses on longer sequences of handwritten scripts. For this reason, we did not explore LSTM memory cells, which are often used to help improve the ability of RNN's to retain information over longer time-scales (usually at a cost in terms of model complexity and computational run-time).

\vspace{-3mm}
\subsubsection{Constrained and unconstrained text recognition}
Recognizing text in the wild without a lexicon or dictionary is a challenging task. Bissacco first reported lexicon-free  (but heavily dictionary weighted) results in \cite{bissa2013}, and Jaderberg \etal \cite{jader2015b} recently presented a notable advancement in lexicon/dictionary-unconstrained scene text recognition. Table \ref{tab:unconstraint_results} compares the accuracy of our proposed method to the previous best results in \cite{jader2015b} on fully unconstrained text recognition task. As can be seen, our method significantly outperforms the JOINT model in \cite{jader2015b} by margin of 9\% on SVT and 8.2\% on ICDAR 2013.

Even though our proposed method aims at the unconstrained scenario, we also compare our results to the constrained setting in which the output is selected with the smallest edit distance between the predicted character sequence and words in the pre-defined lexicon. Table \ref{tab:final_table} shows these comparisons. Our method obtained the new best results for unconstrained recognition on several benchmarks, especially the recent released ICDAR 2013 (IC13) dataset. We also report unconstrained text recognition result on IIIT5k that has not been recorded in previous literature. We are also competitive with the very best results in the constrained setting as well. (Notice that the DICT model of \cite{jader2015} is trained on a specific dictionary that contains ground-truth words for the test set; it is not able to handle previously unseen word strings.)

Figure \ref{fig:examples} demonstrates the lexicon-free scene text recognition results by the proposed R$^2$AM framework on examples from SVT, ICDAR, and IIIT5k datasets. The R$^2$AM method is able to recognize words with low contrast, significant transform and clutter background. It is also able to recover missing/occluded characters by the implicitly learned language model. We further demonstrate our language model by performing text prediction on non-text images as shown in Figure \ref{fig:nontext-examples}, seeing that our method can exploit the underlying character-level statistics and produce word-like strings even though the images have no alphanumeric characters.

\vspace{-2mm}
\section{Conclusion and future directions}
We have presented a new lexicon-free photo OCR framework that incorporates recursive CNNs for image encoding, RNNs for language modeling, and attention-based mechanism for better image feature usage. Extensive analysis shows the effectiveness of each of the proposed component, generalizability to both constrained and unconstrained scenarios with the same methodology, and the practical ability of the proposed method to recognize real-world scene text with state-of-the-art results.

In the future we will explore recursive fully convolutional networks for better connecting extracted image features and corresponding location on input image, and to visualize attention coefficients on the input domain. Also, we will adopt gated units to allow incoming signal to alter the state of attention-based mechanism \cite{lee2016generalizing} and inject deep supervision \cite{lee2015deeply}.

A further future direction is to extend this work into text-detection \cite{yao2012detecting, veit2016coco} in addition to reading, leading to an end-to-end full-image-to-text reading pipeline.

\vspace{-2mm}
\section*{Acknowledgements}
\vspace{-1mm}
The authors thank Jack Culpepper and Cyprien Noel for valuable discussion, and also thank Flickr Vision and Machine Learning Team for supporting the project.

{\footnotesize
\bibliographystyle{ieee}
\bibliography{egbib}
}

\setcounter{section}{0}
\renewcommand{\thesection}{A\arabic{section}}
\setcounter{table}{0}
\renewcommand{\thetable}{A\arabic{table}}
\setcounter{figure}{0}
\renewcommand{\thefigure}{A\arabic{figure}}

\begin{figure*}[t]
\begin{center}
\includegraphics[width=0.9\linewidth]{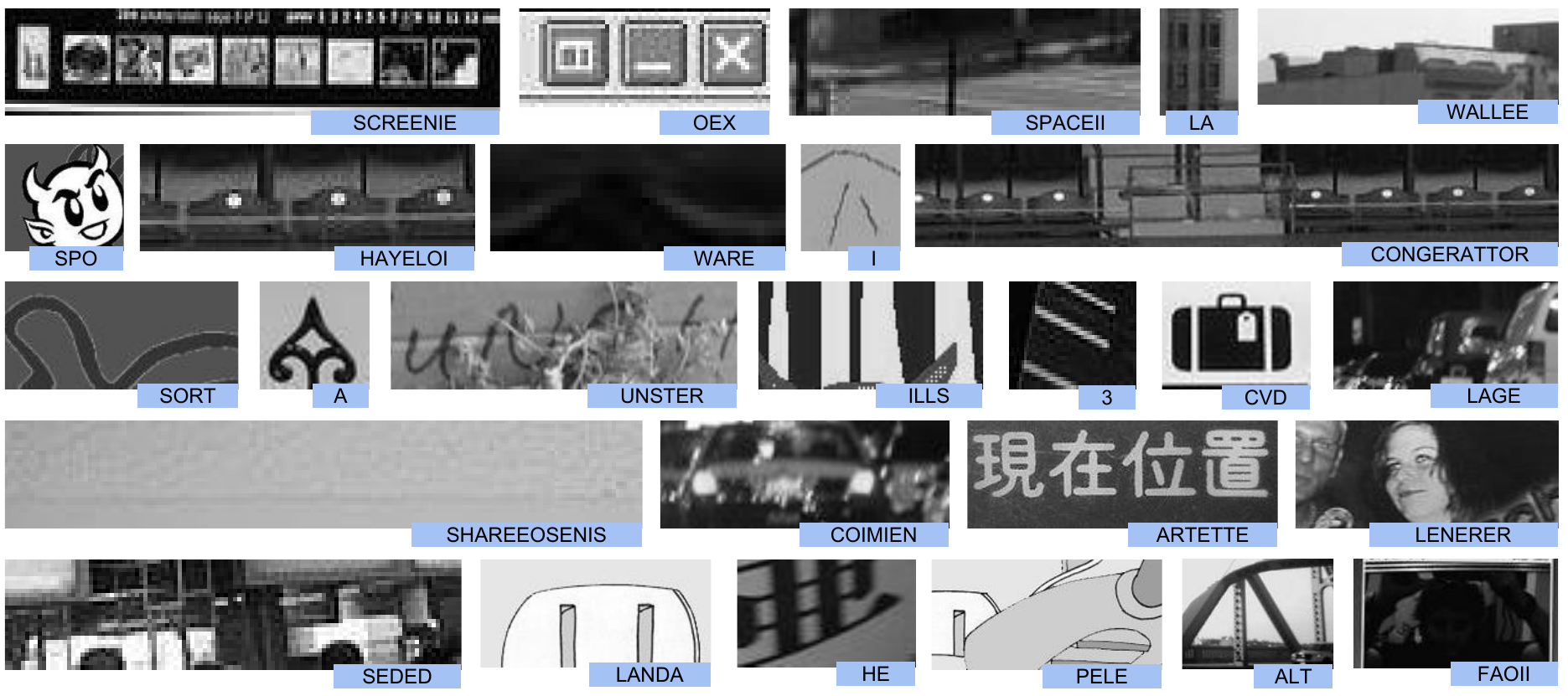}
\end{center}
\vspace{-6mm}
   \caption{Lexicon-free output predictions on non-alphanumeric-text images by the proposed Recursive Recurrent Nets with Attention Modeling (R$^2$AM) framework. By directly operating on images without alphanumeric characters, we can see our model produces output characters that are best fit to the underlying character-level language model implicitly learned from the training data.}
\label{fig:nontext-examples}
\end{figure*}

\definecolor{maroon}{cmyk}{1,1,1,1}

\begin{table*}[!ht]
\small
    \begin{minipage}{.21\linewidth}
      \centering
\begin{tabular}{|c|}
\hline
fc10-4096       \\
fc9-4096        \\
\hline
Conv8-512 3x3                    \\
Conv7-512 3x3                    \\
\hline
Maxpool 2x2 stride:2            \\
\hline
Conv6-256 3x3                    \\
Conv5-256 3x3                    \\
\hline
Maxpool 2x2 stride:2            \\
\hline
Conv4-128 3x3                   \\
Conv3-128 3x3                   \\
\hline
Maxpool 2x2 stride:2            \\
\hline
Conv2-64 3x3                    \\
Conv1-64 3x3                    \\
\hline
\end{tabular}
         \caption*{Base CNN}
    \end{minipage}%
    \begin{minipage}{.21\linewidth}
      \centering
\begin{tabular}{|c|}
\hline
fc10-4096       \\
fc9-4096        \\
\hline
\rowcolor{maroon!2}
Conv8-512 3x3                    \\
\rowcolor{maroon!2}
Conv8-512 3x3                    \\
Conv7-512 3x3                    \\
\hline
Maxpool 2x2 stride:2            \\
\hline
\rowcolor{maroon!2}
Conv6-256 3x3                    \\
\rowcolor{maroon!2}
Conv6-256 3x3                    \\
Conv5-256 3x3                    \\
\hline
Maxpool 2x2 stride:2            \\
\hline
\rowcolor{maroon!2}
Conv4-128 3x3                   \\
\rowcolor{maroon!2}
Conv4-128 3x3                   \\
Conv3-128 3x3                   \\
\hline
Maxpool 2x2 stride:2            \\
\hline
\rowcolor{maroon!2}
Conv2-64 3x3                    \\
\rowcolor{maroon!2}
Conv2-64 3x3                    \\
Conv1-64 3x3                    \\
\hline
\end{tabular}
 \label{tab:recurisveCNN}
         \caption*{Recursive CNN \\ (2 iters)}
    \end{minipage} 
    \begin{minipage}{.21\linewidth}
      \centering
\begin{tabular}{|c|}
\hline
fc10-4096       \\
fc9-4096        \\
\hline
\rowcolor{maroon!2}
Conv8-512 3x3                    \\
\rowcolor{maroon!2}
Conv8-512 3x3                    \\
\rowcolor{maroon!2}
Conv8-512 3x3                    \\
Conv7-512 3x3                    \\
\hline
Maxpool 2x2 stride:2            \\
\hline
\rowcolor{maroon!2}
Conv6-256 3x3                    \\
\rowcolor{maroon!2}
Conv6-256 3x3                    \\
\rowcolor{maroon!2}
Conv6-256 3x3                    \\
Conv5-256 3x3                    \\
\hline
Maxpool 2x2 stride:2            \\
\hline
\rowcolor{maroon!2}
Conv4-128 3x3                   \\
\rowcolor{maroon!2}
Conv4-128 3x3                   \\
\rowcolor{maroon!2}
Conv4-128 3x3                   \\
Conv3-128 3x3                   \\
\hline
Maxpool 2x2 stride:2            \\
\hline
\rowcolor{maroon!2}
Conv2-64 3x3                    \\
\rowcolor{maroon!2}
Conv2-64 3x3                    \\
\rowcolor{maroon!2}
Conv2-64 3x3                    \\
Conv1-64 3x3                    \\
\hline
\end{tabular}
 \label{tab:recurisveCNN}
         \caption*{Recursive CNN \\ (3 iters)}
    \end{minipage} 
    \begin{minipage}{.32\linewidth}
      \centering
\includegraphics[width=1\linewidth]{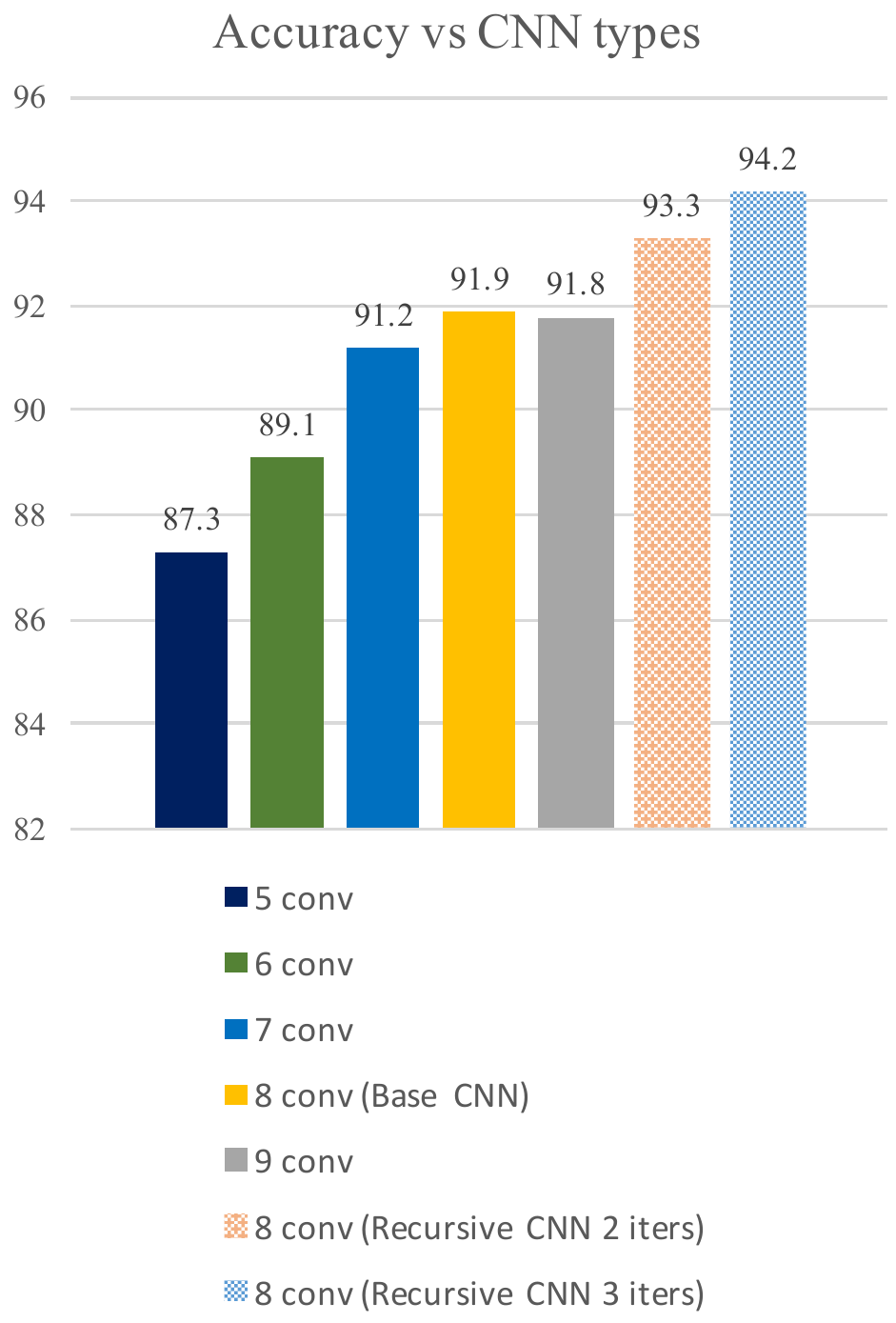}
         \caption*{Performance on Synth90k}
    \end{minipage}%
\vspace{-2mm}
\caption{Left: network architectures for Base CNN and the proposed untied recursive CNNs. Right: the bar chart shows the corresponding performance for networks with different depths on Synth90k dataset. In this experiment we gradually increase the depth of the baseline CHAR model in \cite{jader2015b} from 5 conv layers until we reach the performance plateau at 8 conv layers (denoted as Base CNN as our strong baseline). However, we can further boost the performance by using the proposed untied recursive CNNs. Notice that our recursive CNNs have the same number of parameters as Base CNN but achieve significantly better accuracy.}
\label{tab:baseCNN}
\end{table*}

\end{document}